\documentclass[10pt,twocolumn,letterpaper]{article}

\usepackage[pagenumbers]{cvpr} 
\usepackage{multirow}   
\usepackage{floatrow}
\usepackage[dvipsnames]{xcolor}

\definecolor{cvprblue}{rgb}{0.21,0.49,0.74}
\usepackage[pagebackref,breaklinks,colorlinks,citecolor=cvprblue]{hyperref}

\def\paperID{417} 
\def\confName{CVPR}
\def\confYear{2024}

\title{Spatial-Aware Latent Initialization for Controllable Image Generation}
\author{Wenqiang Sun \and 
Teng Li \and 
Zehong Lin \and 
Jun Zhang \and 
The Hong Kong University of Science and Technology \\
{\tt\small \{wsunap, tliby\}@connect.ust.hk, \{eezhlin, eejzhang\}@ust.hk}
}

\begin{document}

\twocolumn[{
\renewcommand\twocolumn[1][]{#1}
\maketitle
     \centering
     \vspace{-0.25cm}
        \includegraphics[width=0.95\linewidth]{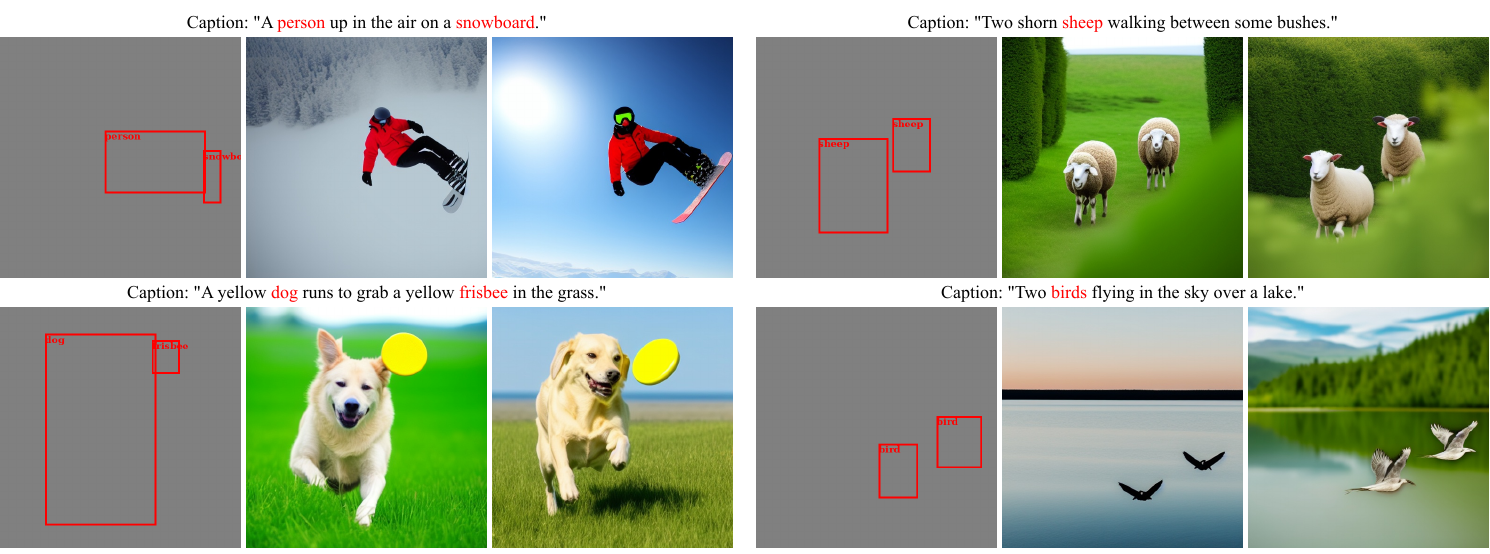}
        \vspace{-0.3cm}
        \captionof{figure}{Our method requires a textual prompt and a bounding box set (the first and forth columns to the left) as the layout condition. The token that has a position requirement is marked in red. For each caption and position pair, we generate images by leveraging our layout guidance approach. As shown in this figure, our method is capable of generating high-quality images with the desired layouts.} \vspace{2em}
        \label{fig:intro}
}]

\begin{abstract}
Recently, text-to-image diffusion models have demonstrated impressive ability to generate high-quality images conditioned on the textual input. However, these models struggle to accurately adhere to textual instructions regarding spatial layout information. While previous research has primarily focused on aligning cross-attention maps with layout conditions, they overlook the impact of the initialization noise on the layout guidance. To achieve better layout control, we propose leveraging a spatial-aware initialization noise during the denoising process. Specifically, we find that the inverted reference image with finite inversion steps contains valuable spatial awareness regarding the object's position, resulting in similar layouts in the generated images. Based on this observation, we develop an open-vocabulary framework to customize a spatial-aware initialization noise for each layout condition. Without modifying other modules except the initialization noise, our approach can be seamlessly integrated as a plug-and-play module within other training-free layout guidance frameworks. We evaluate our approach quantitatively and qualitatively on the available Stable Diffusion model and COCO dataset. Equipped with the spatial-aware latent initialization, our method significantly improves the effectiveness of layout guidance while preserving high-quality content.
\end{abstract}      
\section{Introduction}

\

In recent years, deep generative models, such as Generative Adversarial Networks (GANs) \cite{goodfellow2014generative}, Variational AutoEncoder (VAE) \cite{rezende2014stochastic}, Flow-based Models \cite{dinh2016density}, and Diffusion Models (DMs) \cite{ho2020denoising,dhariwal2021diffusion,rombach2022high,ramesh2022hierarchical,saharia2022photorealistic}, have all demonstrated immense potential in generating realistic images by incorporating extensive training datasets and scalable architectures. Among them, diffusion models have emerged as cutting-edge generative models due to their impressive ability to synthesize high-quality data by gradually denoising the initialization Gaussian noise. In particular, text-to-image diffusion models, such as Stable Diffusion \cite{rombach2022high}, DALL-E-2 \cite{ramesh2022hierarchical}, and Imagen \cite{saharia2022photorealistic}, can generate imaginative images given textual input, dramatically improving the efficiency of artists. 

Despite the significant achievements of current text-to-image diffusion models in image generation, there remain important challenges that need to be addressed, particularly in achieving precise control over the position description provided in the textual prompt. Previous work \cite{gokhale2022benchmarking} evaluates state-of-the-art (SOTA) text-to-image diffusion models and reveals that even these models struggle to comprehend the spatial location requirements within the textual inputs. The primary obstacle lies in the semantic gap between texts and images, indicating that relying solely on texts is insufficient to convey spatial information effectively in images.

To address this issue, previous studies \cite{zheng2023layoutdiffusion,zhang2023layoutdiffusion} propose to train diffusion models from scratch by using layout-image data pairs. While these approaches achieve commendable layout controllability, they overlook the potential of pre-trained text-to-image diffusion models and require a substantial amount of annotated data pairs. Moreover, their generalization capability in open environments is limited \cite{zheng2023layoutdiffusion}. On the other hand, to avoid additional training or fine-tuning of pre-trained diffusion models, recent works \cite{chen2023training,couairon2023zero} propose optimizing the noisy data during the sampling process to align cross-attention maps with layout conditions. Unfortunately, these methods fail to consistently achieve satisfactory performance in terms of both the layout control accuracy and image quality. Different from previous works, Mao \etal \cite{mao2023guided} discover that layout control can be achieved by modifying the initialization noise. 
However, their proposed method can only achieve rough guidance based on the provided spatial description.

Note that the image generation process is primarily influenced by the initialization noise, the diffusion model, and the text prompt. In particular, we empirically find that the initialization noise plays an important role in zero-shot layout guidance approaches \cite{chen2023training,couairon2023zero}.  Specifically, the presence of spatial layout information within the initialization noise helps enhance the layout guidance, and the accuracy is proportional to the level of such spatial awareness. Motivated by this, we propose utilizing a spatial-aware latent that contains spatial layout information within the provided layout area as the initialization noise for sampling images. To further enhance layout control, we introduce an additional attention guidance process during sampling. Notably, due to our spatial-aware latent initialization approach, the number of optimization iterations required is significantly reduced. 
It is worth noting that our proposed method does not require any additional training or fine-tuning on pre-trained text-to-image diffusion models. As illustrated in Fig. \ref{fig:intro}, given textual prompts and layout requirements, our approach achieves controllable generation in terms of the layout and quality of generated images.

In summary, our contributions are as follows:

\begin{itemize}
    \item Through empirical studies, we find that the cross-attention map alignment method achieves effective layout guidance when the layout of images denoised from an initialization noise is close to the target position description. However, random initialization noise lacks such spatial awareness. Consequently, the resultant layout of generated images may be significantly influenced by other factors such as the denoising process and prompts. To address this issue, we aim to find a spatial-aware initialization noise for better layout control.
    \item We propose utilizing the DDIM \cite{song2020denoising} inversion latent from finite steps as the spatial-aware initialization noise. In particular, given various prompts, the spatial information introduced by the DDIM inversion latent remains stable throughout the denoising process. 
    \item Our proposed spatial-aware latent initialization can be seamlessly integrated as a plug-and-play module into existing layout guidance frameworks to enhance the effectiveness of layout control.
    \item We provide extensive experimental results, which demonstrate that our proposed method outperforms SOTA zero-shot layout guidance methods on IoU and mAP@0.5 by more than 15\% and 25\%, respectively, while maintaining a comparable CLIP score.
\end{itemize}

\section{Related Work}
\textbf{Text-to-Image Generation}.\quad For the past few years, GANs \cite{goodfellow2014generative} have dominated the text-to-image generation area \cite{reed2016generative,xu2018attngan,zhu2019dm,li2019controllable}. More recently, diffusion models \cite{ho2020denoising,song2020denoising} have attracted increasing attention due to their impressive generation performance with respect to the stability of the training process and quality of synthetic images. The ablated diffusion model with guidance (ADM-G) \cite{dhariwal2021diffusion} firstly introduces classifier guidance for label-conditional generation and modifies the U-Net architecture in diffusion models, achieving superior performance over state-of-the-art GANs. Afterward, to avoid the extra training of classifiers, classifier-free guidance is proposed \cite{ho2022classifier}. Based on this guidance method, several text-to-image diffusion models \cite{zhang2023text,yang2022diffusion} have been developed to create more realistic and human-style images from text prompts. For instance, LDM \cite{rombach2022high} proposes compressing images into the latent space to accelerate the training and sampling process. Notably, this approach utilizes the cross-attention mechanism to incorporate textual guidance in the image generation. Dall-E-2 \cite{ramesh2022hierarchical} introduces a multimodal latent space to align the text inputs and synthetic images. Moreover, Imagen \cite{saharia2022photorealistic} adopts a pre-trained large language model \cite{raffel2020exploring} as the text encoder to enhance the understanding of text inputs. Despite these advancements, accurately synthesizing images based on position descriptions in text prompts remains a challenge for all text-to-image generative models.

\noindent \textbf{Layout Conditioned Image Generation}.\quad  Recent works \cite{gokhale2022benchmarking,avrahami2023spatext} have highlighted the challenges of guiding object layout generation solely using text inputs. Many existing approaches \cite{park2019semantic,sun2019image,sushko2020you,yang2022modeling} address this challenge by incorporating additional spatial conditions such as bounding boxes or segmentation masks to guide the layout in image generation. These methods, however, do not consider the text inputs and are typically trained on large annotated datasets. Inspired by the layout conditioned generation methods, several studies \cite{wang2022semantic,li2023gligen,zheng2023layoutdiffusion,zhang2023adding,mou2023t2i} have proposed combining text inputs with bounding boxes or segmentation masks to achieve better layout control. For instance, GLIGEN \cite{li2023gligen} introduces an additional attention layer to a pre-trained diffusion model to integrate both text and bounding box inputs. LayoutDiffusion \cite{zheng2023layoutdiffusion} presents a layout-to-image diffusion model that incorporates layout inputs along with text prompts as training data pairs. Moreover, ControlNet \cite{zhang2023adding} and T2I \cite{mou2023t2i} utilize adapters to extend a pre-trained diffusion model with different spatial controls. However, all these approaches require training on large layout annotated datasets, which can be time-consuming and may potentially affect the performance of pre-trained diffusion models.

\noindent \textbf{Training-Free Layout Guidance in Image Generation}.\quad Recent works, prompt-to-prompt \cite{hertz2022prompt} and pix2pix-zero \cite{parmar2023zero}, have highlighted the strong correlation between cross-attention maps and synthetic content in images. Based on this observation, prompt-to-prompt \cite{hertz2022prompt} proposes training-free image editing methods by directly modifying the cross-attention maps. However, these methods primarily focus on local editing of images and lack the ability to control spatial layouts. To achieve training-free layout control, there are studies \cite{chen2023training,couairon2023zero,epstein2023diffusion} leveraging the relationship between cross-attention maps and spatial layout to optimize the sampling process. For instance, Chen \etal \cite{chen2023training} propose optimizing the noisy latent during the sampling process to align cross-attention maps with provided bounding boxes. 
Similarly, fine-grained layout guidance is introduced in \cite{couairon2023zero} by utilizing ground-truth segmentation masks. 
Moreover, Epstein \etal \cite{epstein2023diffusion} demonstrate that cross-attention maps can effectively capture object characteristics such as shape, size, and location in images, enabling various image edits like local editing, resizing, and location movements. 
Nonetheless, these approaches do not always ensure reliable performance in terms of simultaneously meeting position requirements and maintaining the quality of content.

The most related work \cite{mao2023guided} finds that the initialization noise affects the position of generated images. 
To achieve layout guidance, it proposes moving pixels with large cross-attention values into the layout bounding box. This method, however, solely focuses on the random noise without specific spatial information, resulting in unsatisfactory guidance efficiency. 
To leverage the spatial layout information, the DDIM inversion latent is employed in \cite{tumanyan2023plug} to achieve fine-grained image-to-image translation by modifying the spatial features and self-attention blocks. Nonetheless, the goal is to translate an image to another style instead of generating images from textual inputs. Moreover, SDEdit \cite{meng2021sdedit} transfers stroke paintings to real images by adding noise to the input image and denoising it using pre-trained diffusion models. Similarly, this method is mostly used in image translation and does not specifically address the challenges of spatial layout guidance in text-to-image generation tasks. In contrast to these works, our approach primarily centers on utilizing a spatial-aware initialization noise to control the layout during image generation based on textual inputs and position requirements.

\section{Method}

\

In this work, we consider the task of layout-guided text-to-image generation, where realistic images are synthesized by progressively denosing a random initialization noise $z_T$ given textual prompts and bounding box specifications. To accomplish this, users are required to input a text prompt of length $N$, denoted by $\mathcal{P} = \{p_1, p_2, ..., p_N\}$, and a set of $S$ bounding boxes, denoted by $\mathcal{B} = \{b_1, b_2, ..., b_S\}$, where $S \leq N$. Each bounding box $b_i$ is associated with a token subset $\mathcal{P}_i \subseteq \mathcal{P}$. The goal is to generate images within each bounding box $b_i$ for specific tokens $\mathcal{P}_i$ while maintaining high-quality content. 

\begin{figure}[t]
     \centering
        \includegraphics[width=1.0\linewidth]{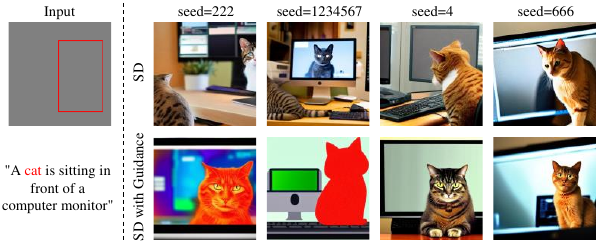}
        \caption{\textbf{Layout Guidance Using Different Seeds.} Given the same random noise, prompt, and bounding box, we generate images using Stable Diffusion (SD) (top row) and SD with the layout guidance method proposed in \cite{chen2023training} (bottom row), respectively. 
        From the same initialization noise $z_T$, when images generated by SD are close to the given layout condition, the layout guidance method achieves effective control and high-quality generation (forth and fifth columns to the left). Otherwise, the layout performance and image quality may degrade (second and third columns to the left). 
        }
        \label{fig:noise}
\end{figure}

We note that the initialization noise plays a critical role in the generation process as it is strongly correlated with the denoising process. In particular, we empirically find that the effectiveness of layout guidance is strongly influenced by the spatial information embedded in the initialization noise. Specifically, as shown in Fig. \ref{fig:noise}, when the ``cat" generated by the initialization noise $z_T$ is closely positioned to the respective bounding box, the layout guidance method enables effective control over the layout. However, if the ``cat" is located far from the bounding box $b_i$, the effectiveness of layout guidance may diminish.
This observation highlights the importance of spatial awareness within the initialization noise in layout guidance.

As demonstrated in \cite{mokady2023null,tumanyan2023plug}, DDIM inversion is an effective technique to capture the spatial information in image translation and editing tasks. Specifically, DDIM inversion serves as the reverse process of the denoising procedure, which involves adding noise for $T$ timesteps from the original latent image $z_0$ to a latent noise $z_T^*$. Moreover, as shown in Fig. \ref{fig:demo_1}, we empirically find that the DDIM inversion latent from an individual image can be generalized to different textual inputs specifying the same position, which implies that it is spatial-aware. 
Therefore, in this paper, we propose to adopt this spatial-aware latent as the initialization noise to achieve effective layout control throughout the generation process.

In Sec. \ref{3.1}, we provide a brief introduction about diffusion models, followed by the analysis about the influence of initialization latent in Sec. \ref{3.2}. Our proposed approach and framework are presented in Sec. \ref{3.3}.

\subsection{Preliminaries} \label{3.1}

\noindent \textbf{Diffusion Models.}\quad Diffusion models consist of two main components: the diffusion process and the denoising process. These models are designed as probabilistic models to learn the underlying data distribution, denoted by $p(x)$. During the diffusion process, Gaussian noises, denoted by $q(x_t|x_{t-1})$, $1 \leq t$, are sequentially added to the input data $x_0$. This process continues until the noisy data approximates a pure Gaussian noise $x_T$. In the subsequent denoising process, the models aim to recover original images $x_0$ by gradually denoising the Gaussian noise $x_T$. Specifically, diffusion models are trained to predict a noise variant, denoted by $p_{\theta}(x_{t-1}|x_t)$, to denoise the noisy data $x_t$, $1 \leq t \leq T$. Following the text-to-image generation process illustrated in LDM \cite{rombach2022high}, the original data $x$ is compressed into latent variables $z$ using an encoder $\mathcal{E}$, such that $z = \mathcal{E}(x)$. With the additional text input $\mathcal{P}$, the simplified objective function for training diffusion models becomes
\begin{equation}
    L_{\text {simple }}(\theta):=\mathbb{E}_{t, z_t, \boldsymbol{\epsilon}, \mathcal{C}}\left[\left\|\boldsymbol{\epsilon}-\boldsymbol{\epsilon}_\theta\left(z_t, t, \mathcal{C}\right)\right\|^2\right],      
\end{equation}
where $z_t$ represents the noisy latent sampled at timestep $t$, $\mathcal{C}=\tau _{\theta}(\mathcal{P})$ is the text embedding, $\tau _{\theta}$ is the text encoder, and $\boldsymbol{\epsilon}$ represents the noise. The noise prediction network $\boldsymbol{\epsilon}_\theta$ is implemented using the U-Net model architecture \cite{ronneberger2015u}. To guide the image generation using textual prompts, we utilize the cross-attention mechanism \cite{vaswani2017attention} as follows:
\begin{equation}
    \operatorname{Attention}(Q, K, V) = M \cdot V ,
\end{equation}
where $M = \operatorname{Softmax}\left(\frac{QK^T}{\sqrt{d}}\right)$ represents the cross-attention map, the query $Q$ is from latent $z$, the key $K$ and value $V$ are computed from the text embedding $\mathcal{C}$, and $d$ is the latent dimension of $Q$ and $K$. Each element $M_{ij}$ in the attention matrix $M$ defines the attention value of the $j$-th token on the pixel patch $i$.

\noindent \textbf{DDIM Inversion.}\quad To perform a deterministic diffusion process from $z_0$ to $z_T$, Denoising Diffusion Implicit Models (DDIMs) \cite{song2020denoising} propose an inversion method as follows:
\begin{equation} \label{DDIM}
\begin{aligned}
    z_{t+1}=&\sqrt{\frac{\alpha_{t+1}}{\alpha_t}} z_t \\
    &+\sqrt{\alpha_{t+1}}\left(\sqrt{\frac{1-\alpha_{t+1}}{\alpha_{t+1}}}-\sqrt{\frac{1-\alpha_t}{\alpha_t}}\right) \cdot \boldsymbol{\epsilon}_\theta\left(z_t, t, \mathcal{C}\right),  
\end{aligned}
\end{equation}
where $z_{t+1}$ denotes noisy latent at timestep $t+1$. 
Note that the added noise $\boldsymbol{\epsilon}_\theta\left(z_t, t, \mathcal{C}\right)$ in timestep $t$ is related to the noisy latent $z_t$, thus establishing the strong relationship between the last latent $z_T$ and the original image latent $z_0$.

\noindent \textbf{Classifier-Free Guidance.}\quad Instead of using a well-trained classifier for guidance as in \cite{dhariwal2021diffusion}, we use classifier-free guidance \cite{ho2022classifier} to guide image generation via interpolation between conditional and unconditional sampling. Specifically, the classifier-free guidance sampling is expressed as:
\begin{equation} \label{guidance_free}
\begin{aligned}
    \hat{\boldsymbol{\epsilon}}_\theta\left(z_t, t, \mathcal{C}, \varnothing\right)=&w \cdot \boldsymbol{\epsilon}_\theta\left(z_t, t, \mathcal{C}\right) \\
    &+(1-w) \cdot \boldsymbol{\epsilon}_\theta\left(z_t, t, \varnothing\right),
\end{aligned}
\end{equation}
where $\varnothing$ denotes the unconditional textual embedding, and $w$ denotes the guidance parameter that balances the conditional and unconditional sampling.

\subsection{The Role of Initialization Noises}  \label{3.2}

\

Inspired by the strong relationship between cross-attention maps and spatial layout of synthetic images, recent works \cite{chen2023training,couairon2023zero} propose to align cross-attention maps with pre-defined bounding boxes or segmentation masks by optimizing the noisy latent in the inference stage. These methods avoid additional training for diffusion models and require low computation cost for online optimization. However, they cannot always provide satisfactory performance with respect to both layout controllability and image quality. A plausible explanation from \cite{chen2023training} is that the randomness of initialization noises may influence the layout guidance process, but the underlying rationale remains unknown.

To fill this gap, we conduct an empirical study to shed light on this issue. As shown in Fig. \ref{fig:noise}, if the synthetic preference of the initialization noise is close to the given bounding boxes, the layout guidance is highly effective. However, starting from a poor initialization noise causes a difficult optimization procedure and low generation quality. This observation emphasizes the importance of initialization noises in the layout control process. Specifically, the spatial awareness contained in the initialization noise strongly affects the effectiveness of layout guidance.

\begin{figure}[!]
     \centering
        \includegraphics[width=0.93\linewidth]{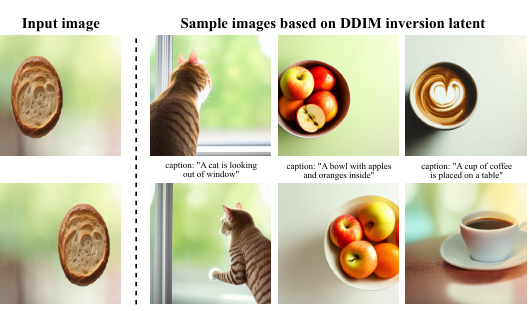}
        \caption{\textbf{Sampling using DDIM inversion latent.} We first invert original images (leftmost column) to obtain DDIM inversion latent $z_T^*$, and use the latent variable $z_T^*$ as the initialization noise to sample images. Given different captions, we clearly see that the layouts of synthetic images (first to third columns to the right) are highly consistent with the input images.}
        \label{fig:demo_1}
\end{figure}

\begin{figure}[!]
     \centering
        \includegraphics[width=0.93\linewidth]{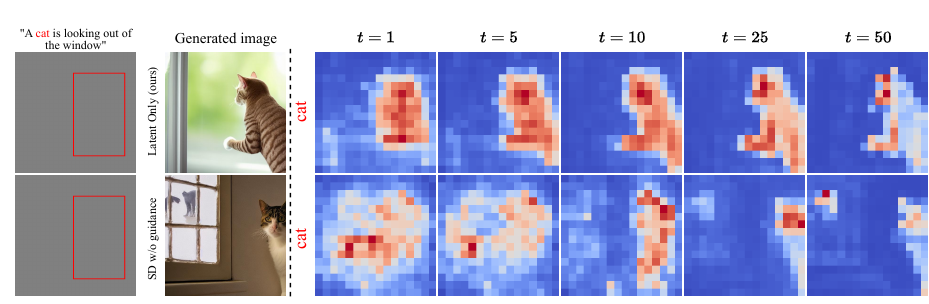}
        \caption{\textbf{Cross-attention map visualization.} By utilizing the DDIM inversion latent $z_T^*$ as the initialization noise (upper row) for image sampling, the cross-attention map of the target object ``cat" shows a consistent tendency across sampling timesteps. In contrast, the cross-attention map changes significantly during sampling when using a random initialization noise (lower row).}
        \label{fig:atten}
\end{figure}

Prior work \cite{mao2023guided} proposes controlling the layout by modifying values of random initialization noise based on their corresponding cross-attention values. However, this method overlooks the significant changes in cross-attention maps of noisy latent variables $z_t$ during the sampling process, leading to unsatisfactory layout controllability. Based on these observations, we propose to find a spatial-aware initialization noise that can maintain consistent spatial position throughout the entire sampling process. Considering the strong relationship between DDIM inversion latent and original data, we conduct experiments to investigate whether the DDIM inversion latent possesses spatial awareness. As shown in Fig. \ref{fig:demo_1}, when using the DDIM inverted reference image latent from finite steps as the initialization noise, the synthetic objects tend to appear in positions that resemble the main object in the reference image. Furthermore, we explore the tendency of position shifts throughout the denoising process. From Fig. \ref{fig:atten}, we observe that the attention values associated with the main object are dominant at the beginning and only undergo slight changes in subsequent steps. Therefore, the DDIM inversion latent is spatial-aware. In addition, as shown in Fig. \ref{fig:demo_1}, we discover that this spatial awareness is independent of the synthetic objects. Even if the synthetic image is totally different from input images in terms of concept and shape, the spatial tendency remains relatively unchanged. This discovery motivates us to utilize the DDIM inversion latent $z_T^*$ as the initialization noise to guide the layout generation.

\subsection{Layout Guidance via Spatial-Aware Latent Initialization}      \label{3.3}

\begin{figure*}[t]
     \centering
        \includegraphics[width=0.92\linewidth]{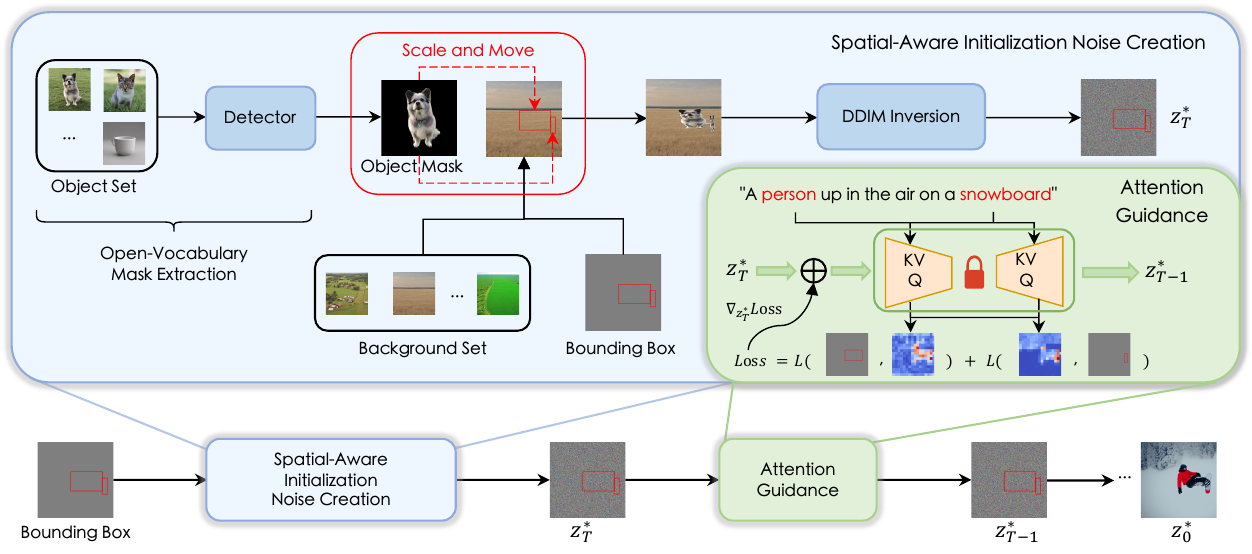}
        \caption{\textbf{Overview of the proposed framework.} Given layout conditions, we customize a reference image by transforming an open-vocabulary object mask to the same shapes and positioning them correspondingly. This image is then processed by DDIM invesion to obtain a spatial-aware latent as the initialization noise. In the denoising process, we utilize the cross-attention map to optimize the latent for more accurate layout control.}
        \label{fig:framework}
\end{figure*}

\noindent \textbf{Creating Spatial-Aware Initialization Noises.}\quad To cater to diverse textual prompts and position requirements, we develop a scalable framework to adapt our method to different scenarios. As shown in Fig. \ref{fig:framework}, we begin with a collected open-vocabulary dataset and employ a pre-trained object detector to extract the object mask $m$ for images from the dataset. Then, we select a background image from a predefined background set. By adhering to the provided layout conditions, we resize and position the object mask $m$ within the background image, resulting in a new image denoted by $I^*$. Following this operation, the positions of the main objects in the new image $I^*$ align with the specified layout requirements. Subsequently, we utilize the DDIM inversion process described in Eq. (\ref{DDIM}) to transform $I^*$ into a spatial-aware latent variable $z_T^*$. Notably, $z_T^*$ effectively captures spatial information within the given bounding boxes. Therefore, instead of initializing the sampling process with a random noise $z_T$, we initiate it with the spatial-aware latent $z_T^*$ to ensure that the target object appears at the desired position. 

In the following, we introduce an additional attention guidance during sampling to enhance layout control.


\noindent \textbf{Attention Guidance.}\quad Inspired by \cite{chen2023training}, we employ cross-attention maps to guide the layout during the inference stage. Initially, attention maps $M_i$ are extracted for each object $\mathcal{P}_i$. The objective is to ensure that high attention values of $M_i$ are predominantly confined within the given bounding box $b_i$. To achieve this, we adopt the following loss function to guide the layout:
\begin{equation}
    L = \sum_{b_i \in \mathcal{B}}\left(\left( 1 - \frac{\sum_{p\sim b_i}M_{pi}}{\sum_{p}M_{pi}} \right)^2 - \lambda \sum_{p\sim b_i}M_{pi}\right), 
\end{equation}
where $M_{pi}$ defines the attention value of object $\mathcal{P}_i$ on the $p$-th pixel in the latent space, and $\lambda$ is a coefficient that balances the alignment and regularization term. Different from the alignment method proposed in \cite{chen2023training}, we introduce a regularization term $\sum_{p\sim b_i}M_{pi}$ to the loss function to ensure that the generated objects are closer to the target positions. For the noisy latent $z_t^*$ at sampling timestep $t$, the loss function $L$ is first used to update the variable $z_t^*$ as follows:
\begin{equation}
    z_t^* = z_t^* - \eta \nabla _{z_t^*}L ,  
\end{equation}
where $\eta$ is the learning rate. Subsequently, $z_{t-1}^*$ is recovered from $z_t^*$ by incorporating the noise $\hat{\boldsymbol{\epsilon}}_{\theta}$ estimated according to Eq. (\ref{guidance_free}). A key distinction between our approach and the one proposed in \cite{chen2023training} lies in the choice of the initialization noise. While \cite{chen2023training} utilizes a random noise $z_T$ as the starting point, we begin with the spatial-aware latent $z_T^*$. 
Note that this distinction helps accelerate the optimization process. Specifically, the method in \cite{chen2023training} requires 10 steps to achieve the desired layout control performance, whereas our approach typically achieves better results in just 1-3 steps, demonstrating superior efficiency in layout control.

\noindent \textbf{Plug-and-Play Module in Mask Guidance.}\quad The proposed spatial-aware initialization latent can be extended to other training-free layout guidance approaches. For instance, ZestGuide \cite{parmar2023zero} proposes leveraging segmentation masks as the layout condition to control the position. This approach is similar to \cite{chen2023training} in aiming to control the layout by aligning the cross-attention maps with the layout condition. Both of them face the challenges associated with the influence of initialization noise on spatial control performance. Thus, we can incorporate our spatial-aware latent initialization as a plug-and-play module into their frameworks.

\section{Experiment}

\

In Sec. \ref{4.1}, we introduce our experimental details, including pre-trained models and dataset. We present our main results in Sec. \ref{4.2} and ablation studies in Sec. \ref{4.3}.
\begin{figure*}[t]
     \centering
        \includegraphics[width=0.935\linewidth]{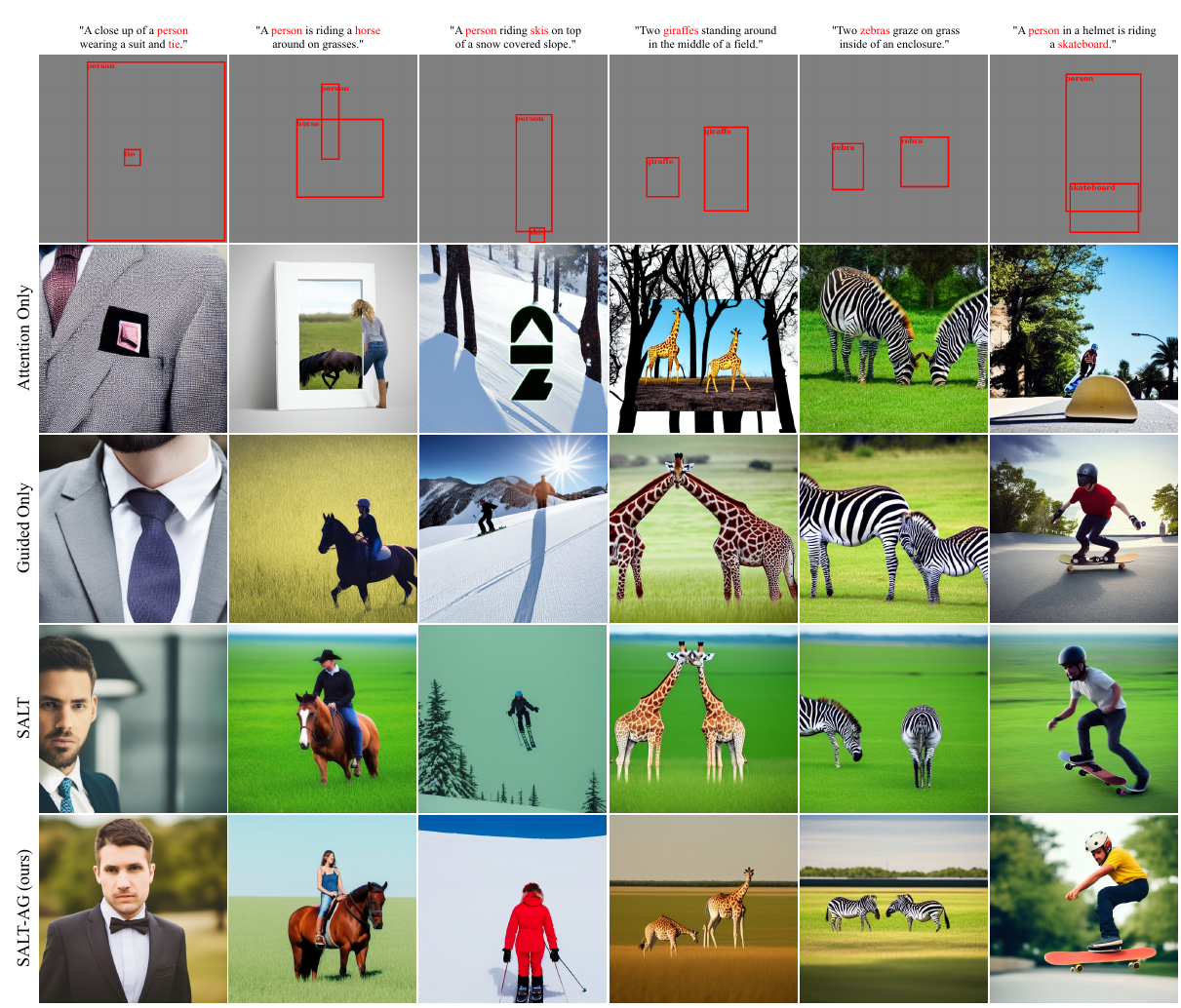}
        \caption{\textbf{Qualitative results for different layout guidance methods.} The generated images are all based on the captions and bounding boxes from COCO. Compared to other guidance methods, our approach achieves the best layout control ability.}
        \label{fig:visualization}
\end{figure*}
\subsection{Experimental Details}\label{4.1}
\textbf{Dataset.}\quad We evaluate the layout guidance efficiency of different methods using captions and bounding boxes from a public dataset, COCO \cite{lin2014microsoft}. Since most images in COCO contain either one or multiple main objects, we first collect captions and detection bounding boxes of 4,040 single-object images to form an evaluation set named COCO-single. Then, we collect captions and bounding boxes of around 800 two-object images to construct another evaluation set called COCO-multiple. We generate one image for each caption and compare the prediction bounding boxes in synthetic images with the target ones. 

\noindent \textbf{Evaluation Metrics.}\quad We use Intersection over Union (IoU) and mean Average Precision (mAP) to measure the layout guidance performance. In our experiments, we set the IoU metric threshold to 0.5 to compute the mAP@0.5. Moreover, we compute the CLIP score \cite{radford2021learning}, which assesses the alignment between generated images and text prompts, to evaluate the quality of synthetic images. 

\noindent \textbf{Baselines.}\quad We compare our method with SOTA training-free guidance baselines. Specifically, \textit{SD} refers to the original Stable Diffusion without layout guidance. \textit{Attention Only} is the approach proposed in \cite{chen2023training}, where the latent is optimized for 10 steps. \textit{Guided Only} \cite{mao2023guided} modifies the value of random initialization noise directly to control the layout. \textit{Attention w/ Guided} is an approach that combines \textit{Attention Only} and \textit{Guided Only}, where the number of optimization steps is set to 10 for fairness. In particular, \textit{SALT} refers to the proposed \textbf{S}patial-\textbf{A}ware \textbf{L}atent ini\textbf{T}ialization, which only modifies the initialization noise in SD, and \textit{SALT-AG} is our proposed method that involves both \textit{SALT} and \textbf{A}ttention \textbf{G}uidance with only 3 steps. 

\noindent \textbf{Implementation Details.}\quad In all experiments, we use Stable Diffusion (SD) V1.5 \cite{rombach2022high} as the pre-trained text-to-image diffusion model. We set both the DDIM inversion and sampling steps to 50. For the DDIM inversion process, we employ the unconditional sampling method. We use Detic \cite{zhou2022detecting} to detect the layout of target objects, and calculate IoU and mAP based on the ground truth bounding box. To compute the CLIP score, we employ the ViT-B/32 provided in \cite{radford2021learning}. Image and text encoders are utilized to extract respective features, and the cosine similarity between these feature vectors is computed as the CLIP score. Note that all images used in our work, including reference objects and background sets, are generated by Stable Diffusion.

\subsection{Main Results} \label{4.2}

\

In this experiment, we use the same reference object of cat and background of green plain for fairness. Ablation studies on different objects and backgrounds are discussed in Sec. \ref{4.3}.

As shown in Table \ref{tb:main result}, our approach \textit{SALT-AG} achieves the best performance in IoU and mAP@0.5 while maintaining a competitive CLIP score across all evaluation settings. In general, the performance of all guidance methods on COCO-multiple is relatively lower than COCO-single since multiple layout conditions pose more challenges. The performance gap between \textit{SALT} and \textit{Guided Only} shows that DDIM inversion latent contains spatial information, whereas the random initialization noise lacks spatial awareness. Compared with \textit{SD}, \textit{SALT} achieves more effective layout control by using DDIM inversion latent as the initialization noise. In addition, our method \textit{SALT-AG} surpasses both \textit{Attention Only} and \textit{Attention w/ Guided} by more than 15\% in IoU and 25\% in mAP@0.5 with much fewer optimization steps. 
These results highlight the crucial role of spatial-aware initialization noise in effective layout control.  

We also provide some qualitative results in Fig. \ref{fig:visualization} to illustrate that our proposed method achieves impressive layout control over the text-to-image generation. In particular, the generated content based on our approach are much more natural and realistic, demonstrating the importance of spatial-aware initialization noise. Specifically, our approach can deal with the challenging layout conditions. For instance, in the the synthetic images corresponding to the prompt ``Two zebras graze on grass inside of an enclosure." (second to the right), our approach successfully adjusts the size and shape of zebras to fit the bounding box, while other methods fail. Moreover, our method can provide satisfactory guidance results to meet the layout requirement for tiny objects. Since the layout guidance effectiveness of \textit{Attention w/ Guided} is similar to that of \textit{Attention Only}, we only present the results of \textit{Attention Only}. Note that the baseline \textit{SD} does not consider the layout condition. 


\begin{table}[t]
\small

\caption{Comparison with other baselines in two evaluation sets. The best results under each metric are marked in bold.
}
\label{tb:main result}
\centering
\resizebox{\linewidth}{!}{
    \begin{tabular}{*{7}{c}}
      \toprule
      \multirow{2}*{Methods} & \multicolumn{3}{c}{COCO-single} & \multicolumn{3}{c}{COCO-multiple} \\
      \cmidrule(lr){2-4}  \cmidrule(lr){5-7}
      & {IoU $\uparrow$} & {mAP@0.5} $\uparrow$ & {CLIP} $\uparrow$& {IoU} $\uparrow$& {mAP@0.5} $\uparrow$& {CLIP} $\uparrow$\\
      \midrule
      SD & 0.19 & 0.13  & 31.28 & 0.20 & 0.13 & \textbf{31.41} \\
      Attention Only \cite{chen2023training} & 0.39 & 0.41 &  30.50 & 0.34 & 0.31 & 30.95 \\
      Guided Only \cite{mao2023guided} & 0.21 & 0.14 &  \textbf{31.36} & 0.21 & 0.15 & 31.27 \\
      Attention w/ Guided & 0.40 & 0.41 &  30.59 & 0.35 & 0.32 & 30.90 \\
      SALT & 0.30 & 0.28 & 30.79 & 0.30 & 0.27 & 30.49 \\
      SALT-AG (ours) & \textbf{0.47} & \textbf{0.54}   & 30.46  & \textbf{0.42} & \textbf{0.43} &  30.42 \\
      \bottomrule
    \end{tabular}
}
\end{table}
\subsection{Ablation Studies} \label{4.3}
\noindent \textbf{The impact of reference object.}\quad We evaluate the performance of our method under different reference objects, including dog, bread and cat. As shown in Table \ref{tab:object}, the variation in IoU and mAP@0.5 results is marginal across all these objects. This demonstrates that the spatial awareness in DDIM inversion latent is object-agnostic.
\begin{table}[t]
\tiny
 \caption{Ablation study for reference object. We evaluate our proposed method on multiple layout condition dataset using three reference objects with the same background, ``gray plain".}
 \label{tab:object}
\centering
\scalebox{1.5}{
    \begin{tabular}{*{4}{c}}
      \toprule
      \multirow{2}*{Object Category} & \multicolumn{3}{c}{COCO-multiple} \\
      \cmidrule(lr){2-4}
      & {IoU $\uparrow$} & {mAP@0.5} $\uparrow$ & {CLIP $\uparrow$}  \\
      \midrule
      bread & 0.40 & 0.40 &  30.79 \\
      cat & 0.38 & 0.39 &  30.72 \\
      dog & 0.39 & 0.38 &  30.60 \\
      \bottomrule
    \end{tabular}
}
\end{table}

\noindent \textbf{The impact of background.}\quad  We analyze the impact of different backgrounds, including the color and content complexity of backgrounds. As shown in Table \ref{tab:background}, the choice of background significantly influences both the layout guidance and generation quality. Deeper color backgrounds tend to result in lower generation quality. In terms of background complexity, the existence of multiple objects in the background leads to worse layout guidance outcomes. This may be because that the additional objects diminish the spatial awareness within the given bounding box. 
\begin{table}[t]
\tiny
 \caption{Ablation study for background. Take object ``dog" as example, different backgrounds are chosen to evaluate the layout control effectiveness. Background ``farm" is the most complex one, ``green plain" is dark background, and ``gray plain" is intermediate background in terms of color and complexity.}
 \label{tab:background}
\centering
\scalebox{1.5}{
    \begin{tabular}{*{4}{c}}
      \toprule
      \multirow{2}*{Background} & \multicolumn{3}{c}{COCO-multiple} \\
      \cmidrule(lr){2-4}
      & {IoU $\uparrow$} & {mAP@0.5} $\uparrow$ & {CLIP $\uparrow$}  \\
      \midrule
      green plain & 0.41 & 0.45 & 30.29  \\
      gray plain & 0.39 & 0.38 & 30.60  \\
      farm & 0.39 & 0.40 & 30.67  \\
      \bottomrule
    \end{tabular}
}
\end{table}
\subsection{Limitations}
\

We note that our method also faces the prompt following challenge. In fact, some works \cite{yu2022scaling} have identified this problem in current SOTA text-to-image generative models. As shown in Table \ref{tb:main result}, in comparison to \textit{SD}, most layout guidance methods result in a decrease of CLIP score. For our proposed method, this challenge may be exacerbated due to the sparsification of spatial awareness in the initialization noise. Specifically, only the area inside the bounding box contains spatial-aware content, which may block the generation guidance from other tokens. Consequently, the CLIP score of our method is slightly lower than the \textit{Attention Only} approach.
\section{Conclusion}

\

In this paper, we analyze the impact of the initialization noise on training-free layout guidance text-to-image generation tasks and highlight the importance of finding a spatial-aware initialization noise. Through empirical studies, we find that the DDIM inversion latent from finite steps contains spatial awareness and thus can be adopted as the initialization noise to achieve effective layout control. Based on these observations, we propose combining the spatial-aware latent initialization and attention guidance to achieve more precise layout control. Evaluation results demonstrate that our method achieves accurate position control while maintaining high-quality image generation. Despite the achievements, there remains a problem shared by most layout guidance approaches. Specifically, these methods tend to neglect the generation of untargeted tokens that do not have specific position requirements. Consequently, the layout guidance may impair the alignment between generated images and prompts. Future works are expected to explore how to balance the layout guidance and image generation given textual prompts and layout conditions.

{
    \small
    \bibliographystyle{ieeenat_fullname}
    \bibliography{main}
}

\clearpage
\maketitlesupplementary

\section{Implementation Details}

\

In the following, we provide additional details about our experiments.

\begin{figure}[!htbp]
     \centering
        \includegraphics[width=1.0\linewidth]{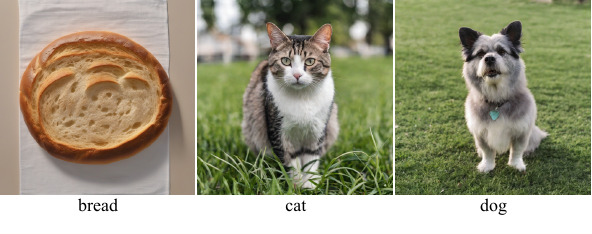}
        \vspace{-0.3cm}
        \caption{Reference objects. We  use three reference objects to evaluate our approach.}
        \label{fig:reference object}
\end{figure}
\begin{figure}[!htbp]
     \centering
        \includegraphics[width=1.0\linewidth]{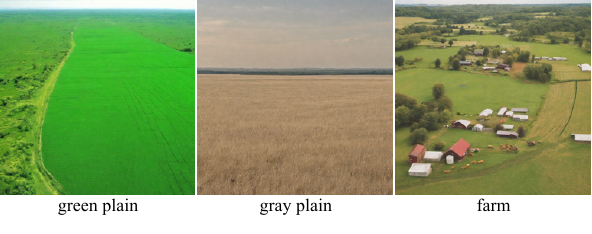}
        \caption{Background set. We  use three backgrounds to evaluate our approach.}
        \label{fig:background}
\end{figure}

\noindent \textbf{Creating Spatial-Aware Initialization Noises.}\quad The reference objects and backgrounds used in our experiments are presented in Fig. \ref{fig:reference object} and Fig. \ref{fig:background}, respectively. In our main results, we use ``cat" and ``green plain" as the reference object and background, respectively, while the other ones are used for ablation studies.


\noindent \textbf{Attention Guidance.}\quad Following the setting in \textit{Attention Only} \cite{chen2023training}, we combine the mid-block and first cross-attention block in up-sampling branch of the U-Net as the ground-truth cross-attention maps to guide the layout. In baseline methods, the attention guidance step is set to 10 for fairness. However, in our proposed approach, we reduce the attention guidance steps to 3 to accelerate the attention guidance process while maintaining comparable layout control performance. The impact of different attention guidance steps will be discussed in Table \ref{tab:guidance_step}. In our experiments, the coefficient $\lambda$ of the regularization term is set to 0.05. 

\noindent \textbf{Baselines.}\quad Since the implementation of the baseline method \textit{Guided Only} \cite{mao2023guided} is not publicly available, we implement this method by ourselves based on the descriptions provided in \cite{mao2023guided}. Specifically, we begin by extracting the corresponding cross-attention maps resulting from the random initialization noise. Based on the given bounding box, we then move the lower-valued pixels outside the bounding box, while moving the higher-valued pixels into the bounding box.  

\section{Ablation Studies}
\noindent \textbf{The impact of attention guidance steps.}\quad We investigate the impact of attention guidance steps on the layout guidance process. Note that the IoU and mAP@0.5 values of the baseline \textit{Attention Only} \cite{chen2023training} are 0.34 and 0.31, respectively. The results in Table \ref{tab:guidance_step} demonstrate that our proposed method achieves better layout performance than the baseline \textit{Attention Only} in terms of both IoU and mAP@0.5, even with just a single attention guidance step. 
To trade off the layout guidance performance, computation efficiency and image quality, we set $\text{steps}=3$ in our proposed approach.


\begin{table}[!htbp]
\tiny
 \caption{Ablation study for attention guidance steps. We evaluate our method on the COCO-multiple set. The ``cat" and ``green plain" serve as the reference object and background, respectively.}
 \label{tab:guidance_step}
\centering
\resizebox{\linewidth}{!}{
    \begin{tabular}{cccccc}
    \toprule
     & steps=1     & steps=2     & steps=3     & steps=5     & steps=10 \\
    \midrule
    IoU $\uparrow$   & 0.40  & 0.41  & 0.42  & 0.43  & 0.41   \\
    mAP@0.5 $\uparrow$ & 0.41  & 0.43  & 0.43  & 0.46  & 0.45  \\
    CLIP $\uparrow$  & 30.45 & 30.21 & 30.42 & 30.17 & 29.78  \\
    \bottomrule
    \end{tabular}}%
  \label{tab:inversion step}%
\end{table}%

\noindent \textbf{The impact of DDIM inversion steps.}\quad We evaluate the layout guidance effectiveness of our method with different numbers of DDIM inversion steps. As shown in Table \ref{tab:inversion step}, a higher number of DDIM inversion steps leads to improved layout control. To reduce the time consumption while maintaining the image quality, we set the number of DDIM inversion steps to 50 in our main results.
\begin{table}[!htbp]
\tiny
 \caption{Ablation study for DDIM inversion steps. We evaluate our approach with different numbers of DDIM inversion steps on COCO-multiple set. For our approach, the reference object is ``cat" and the background is ``green plain".}
 \label{tab:inversion step}
\centering
\scalebox{1.5}{
    \begin{tabular}{*{4}{c}}
      \toprule
      \multirow{2}*{Inversion Steps} & \multicolumn{3}{c}{COCO-multiple} \\
      \cmidrule(lr){2-4}
      & {IoU $\uparrow$} & {mAP@0.5} $\uparrow$ & {CLIP $\uparrow$}  \\
      \midrule
      20 & 0.40 & 0.42 & 29.56  \\
      50 & 0.42 & 0.43 & 30.42  \\
      100 & 0.42 & 0.45 & 30.27  \\
      \bottomrule
    \end{tabular}
}
\end{table}



\noindent \textbf{The impact of the random initialization noise.}\quad To study the impact of the random initialization noise on the layout guidance, we evaluate the performance variation of the baseline method \textit{Attention Only} under different random seeds. The results in Table \ref{tab:noise} show that the variation of the layout guidance performance is extremely marginal. This implies that the layout guidance is robust to the random initialization noise. Therefore, we choose a fixed random seed for the baselines in our experiments.

\begin{table}[!htbp]
\tiny
 \caption{Ablation study for the random initialization noise. We provide the mean and variance of layout guidance metrics based on the \textit{Attention Only} method.}
 \label{tab:noise}
\centering
\resizebox{\linewidth}{!}{
    \begin{tabular}{*{4}{c}}
      \toprule
      \multirow{2}*{Methods} & \multicolumn{3}{c}{COCO-single} \\
      \cmidrule(lr){2-4} 
      & {IoU $\uparrow$} & {mAP@0.5} $\uparrow$ & {CLIP $\uparrow$} \\
      \midrule
      SD & 0.19 $\pm$ 0.01 & 0.13 $\pm$ 0.01 & 31.28 $\pm$ 0.10  \\
      Attention Only \cite{chen2023training} & 0.39 $\pm$ 0.01 & 0.41 $\pm$ 0.01 & 30.50 $\pm$ 0.05  \\
      \bottomrule
    \end{tabular}
}
\end{table}

\noindent \textbf{The impact of regularization term.}\quad We evaluate the impact of our proposed regularization term in the attention guidance module. As shown in Table \ref{tab:regularization}, this term helps achieve a better layout control, in terms of IoU and mAP@0.5, while maintaining a competitive CLIP score for the baseline and our proposed approach. In the main results, our approach adopts the regularization term to enhance the layout guidance.
\begin{table}[!htbp]
\tiny
 \caption{Ablation study for regularization term.}
 \label{tab:regularization}
\centering
\resizebox{\linewidth}{!}{
    \begin{tabular}{*{4}{c}}
      \toprule
      \multirow{2}*{Methods} & \multicolumn{3}{c}{COCO-multiple} \\
      \cmidrule(lr){2-4}
      & {IoU $\uparrow$} & {mAP@0.5} $\uparrow$ & {CLIP $\uparrow$}  \\
      \midrule
      Attention Only w/o regularization \cite{chen2023training} & 0.34 & 0.31 & 30.95  \\
      Attention Only w/ regularization & 0.36 & 0.34 & 30.78  \\
      SALT-AG w/o regularization & 0.41 & 0.42 & 30.47  \\
      SALT-AG w/ regularization (ours) & 0.42 & 0.43 & 30.42   \\
      \bottomrule
    \end{tabular}
}
\end{table}

\noindent \textbf{The choice of spatial-aware initialization noise.}\quad In general, SDE or ODE are two optional formulations to add noises to clean data. 
To demonstrate the necessity of utilizing ODE formulation, we compare our method with SDEdit \cite{meng2021sdedit}.
Specifically, we combine the SDEdit with additional attention guidance, called \textit{SDEdit w/ Attention}. As shown in Table \ref{tab:SDEdit hybrid},
our approach \textit{SALT-AG} with DDIM inversion latent can achieve better layout control while maintaining high-quality generation.
This is because the non-deterministic noise in SDE formulation will destroy the spatial awareness in the latent. 
Therefore, in our approach, 
we adopt the DDIM inversion latent as the spatial-aware initialization noise.


\begin{table}[!htbp]
\tiny
 \caption{Comparison with SDE noisy latent. We evaluate the layout control effectiveness of SDEdit w/ Attention on COCO-multiple set. The reference object is ``cat" and the background is ``green plain".}
 \label{tab:SDEdit hybrid}
\centering
\resizebox{\linewidth}{!}{
    \begin{tabular}{*{4}{c}}
      \toprule
      \multirow{2}*{Methods} & \multicolumn{3}{c}{COCO-multiple} \\
      \cmidrule(lr){2-4}
      & {IoU $\uparrow$} & {mAP@0.5} $\uparrow$ & {CLIP $\uparrow$}  \\
      \midrule
      SDEdit w/ Attention (steps = 3) & 0.36 & 0.34 & 30.73  \\
      SDEdit w/ Attention (steps = 10) & 0.37 & 0.37 & 30.12  \\
      SALT-AG (ours) (steps = 3) & 0.42 & 0.43 & 30.42   \\
      \bottomrule
    \end{tabular}
}
\end{table}

\noindent \textbf{The impact of pure white background.}\quad We adopt the pure white background to exclude the impact of backgrounds on the layout guidance. Note that, based on the ``green plain" background, the IoU, mAP@0.5 and CLIP score are 0.47, 0.54, and 30.46, respectively, using our layout guidance approach.  As shown in Table \ref{tab:white} and Fig.~\ref{fig:white_background}, in comparison to the performance based on other backgrounds in the main result, the pure white background severely impairs 
the generation quality. One potential explanation is that the pure white background almost does not exist in the training dataset of diffusion models, resulting in the low effectiveness of DDIM inversion latent variable.

\begin{table}[!htbp]
\tiny
 \caption{Evaluation results of pure white background.}
 \label{tab:white}
\centering
\scalebox{1.5}{
    \begin{tabular}{*{4}{c}}
      \toprule
      \multirow{2}*{Object Category} & \multicolumn{3}{c}{COCO-single} \\
      \cmidrule(lr){2-4}
      & {IoU $\uparrow$} & {mAP@0.5} $\uparrow$ & {CLIP $\uparrow$}  \\
      \midrule
      bread & 0.41 & 0.45 & 26.14  \\
      cat & 0.41 & 0.47 & 26.71  \\
      dog & 0.42 & 0.47 & 26.74  \\
      \bottomrule
    \end{tabular}
}
\end{table}

\section{Additional Qualitative Results}

\begin{figure*}[!htbp]
     \centering
        \includegraphics[width=1.0\linewidth]{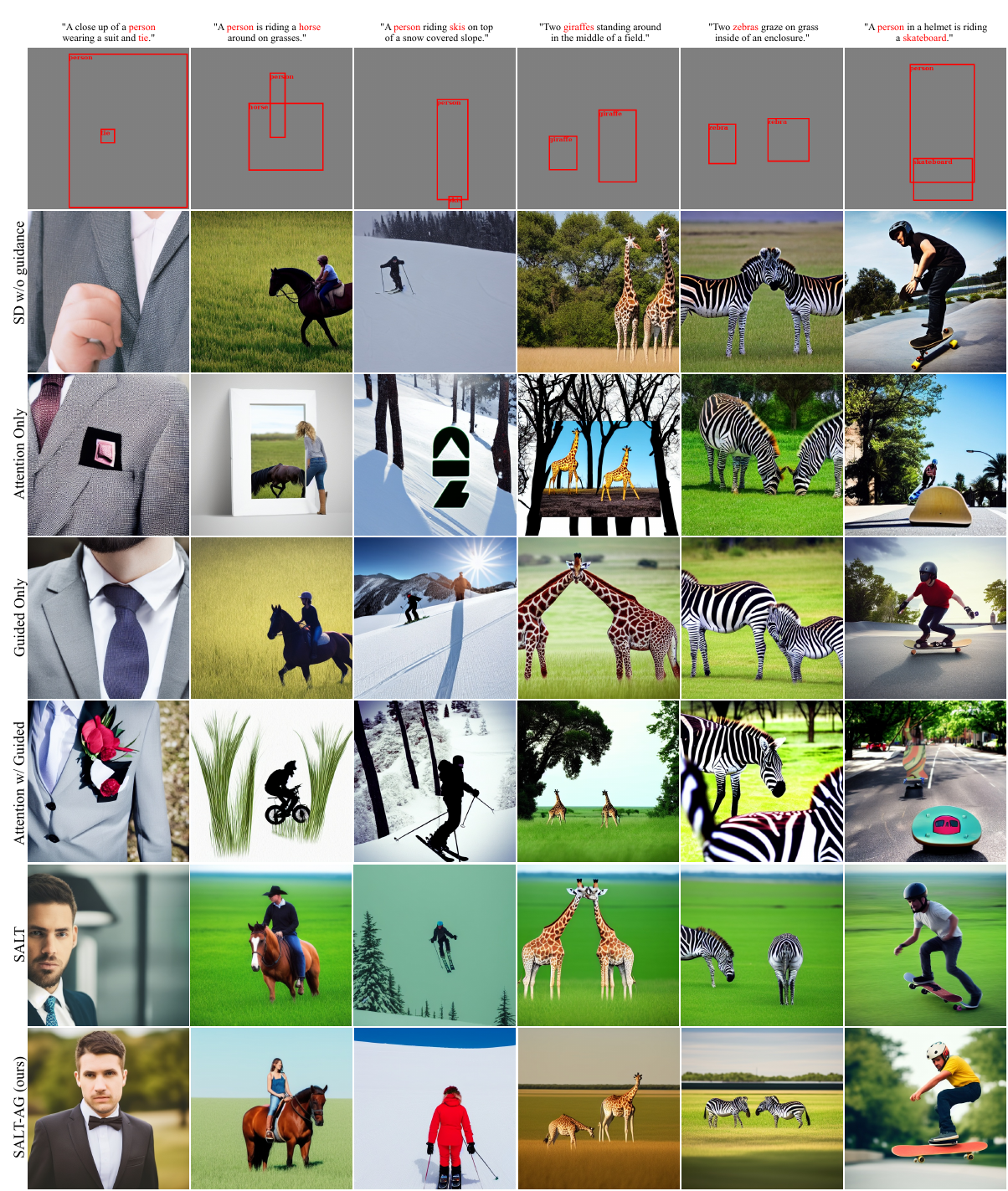}
        \caption{\textbf{Qualitative results for different layout guidance methods.} Generated images are all based on the given captions and bounding boxes. Compared with other baselines, our approach achieves the best layout control while maintaining the high-quality generation.}
        \label{fig:main result}
\end{figure*}

\noindent \textbf{Main results.}\quad We provide the qualitative results for all the baselines and our approach. As shown in Fig. \ref{fig:main result}, our approach outperforms other baselines in terms of layout guidance effectiveness and generation quality.

\begin{figure*}[!htbp]
     \centering
        \includegraphics[width=1.0\linewidth]{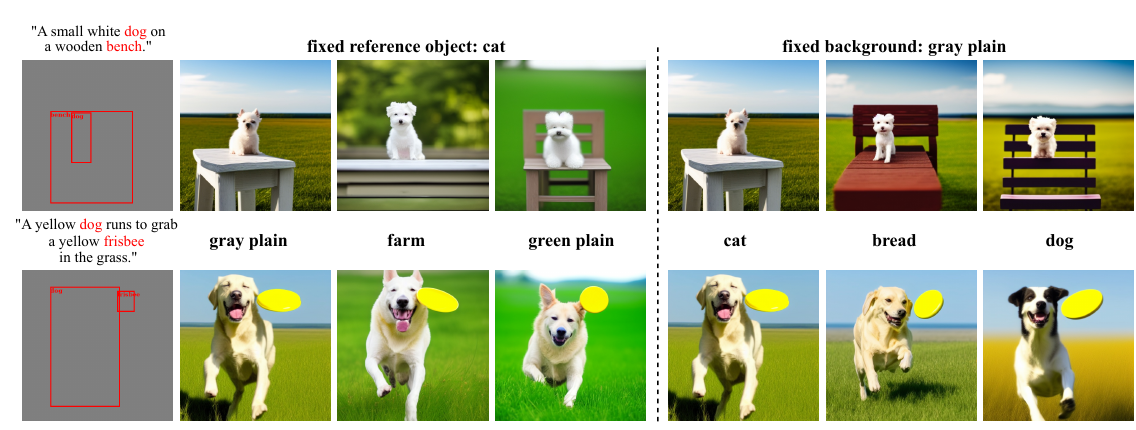}
        \caption{\textbf{Qualitative results for different reference objects and backgrounds.} Given a fixed reference object ``cat", we provide the generated images for different backgrounds (second to forth columns to the left). Moreover, we also present the results for different objects using a fixed background ``gray plain" (fifth to seventh columns to the left).}
        \label{fig:reference objects}
\end{figure*}

\begin{figure*}[!htbp]
     \centering
        \includegraphics[width=1.0\linewidth]{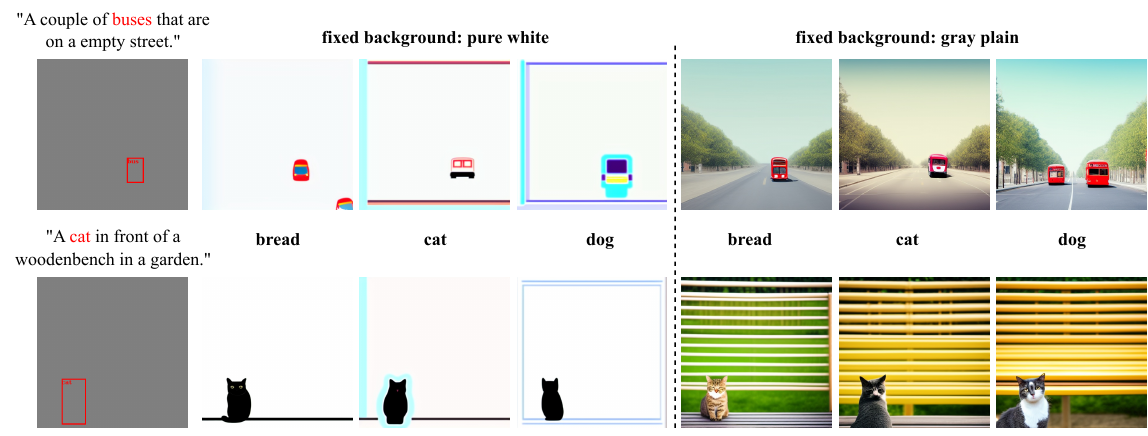}
        \caption{\textbf{Qualitative results for pure white and other backgrounds.} Given the same prompts, bounding box, and reference object, we present the results based on the pure white background (second to forth columns to the left). Moreover, we provide the generated images based on the gray plain background (fifth to seventh columns to the left).}
        \label{fig:white_background}
\end{figure*}

\noindent \textbf{Evaluation results of different reference objects and backgrounds.}\quad As shown in Fig. \ref{fig:reference objects}, we provide the synthetic images generated by our proposed method with different reference objects and backgrounds. It is clear that the background significantly influences the image generation process. Moreover, the evaluation result of pure white background is presented in Fig. \ref{fig:white_background}, which demonstrates that the use of pure white background is harmful to the 
generation quality.


\begin{figure*}[t]
     \centering
        \includegraphics[width=1.0\linewidth]{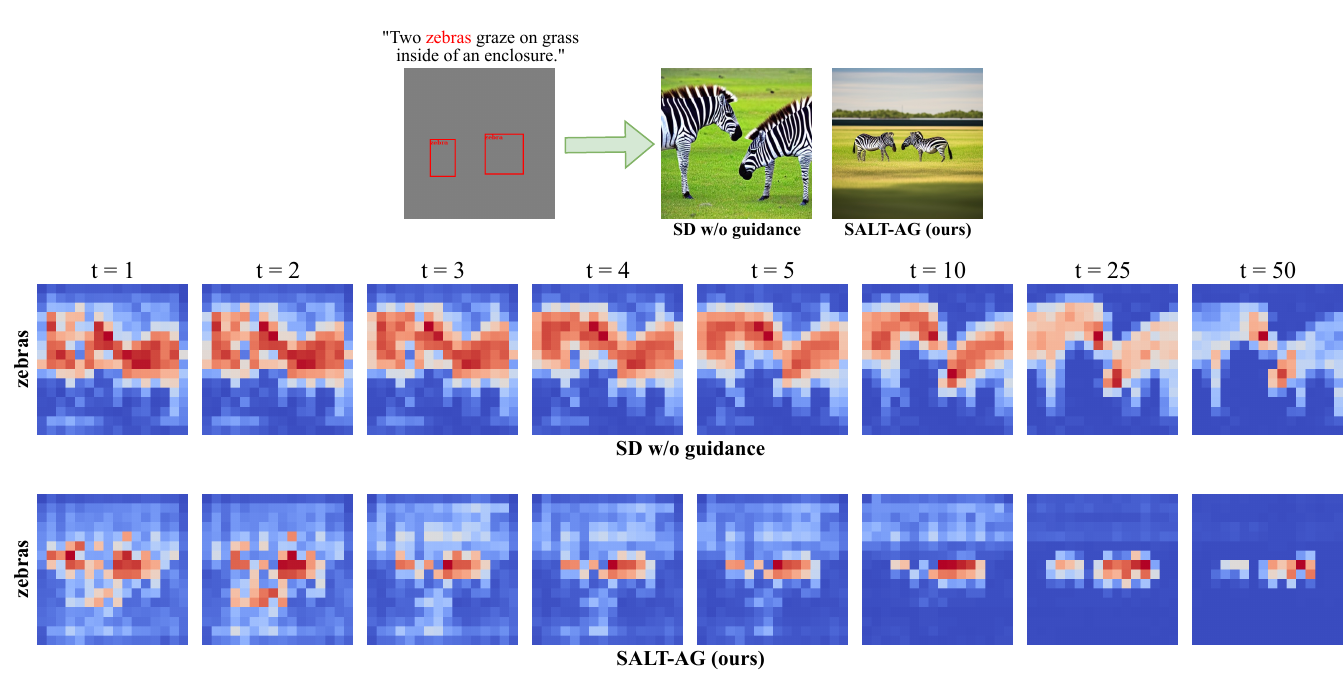}
        \caption{\textbf{Visualization of cross-attention maps during the denoising process.} Based on the given textual prompts and bounding boxes, we present the cross-attention maps of SD w/o guidance (second row to the top) and our layout guidance approach (third row to the top) across different denoising steps.}
        \label{fig:append_atten_1}
\end{figure*}

\begin{figure*}[!htbp]
     \centering
        \includegraphics[width=1.0\linewidth]{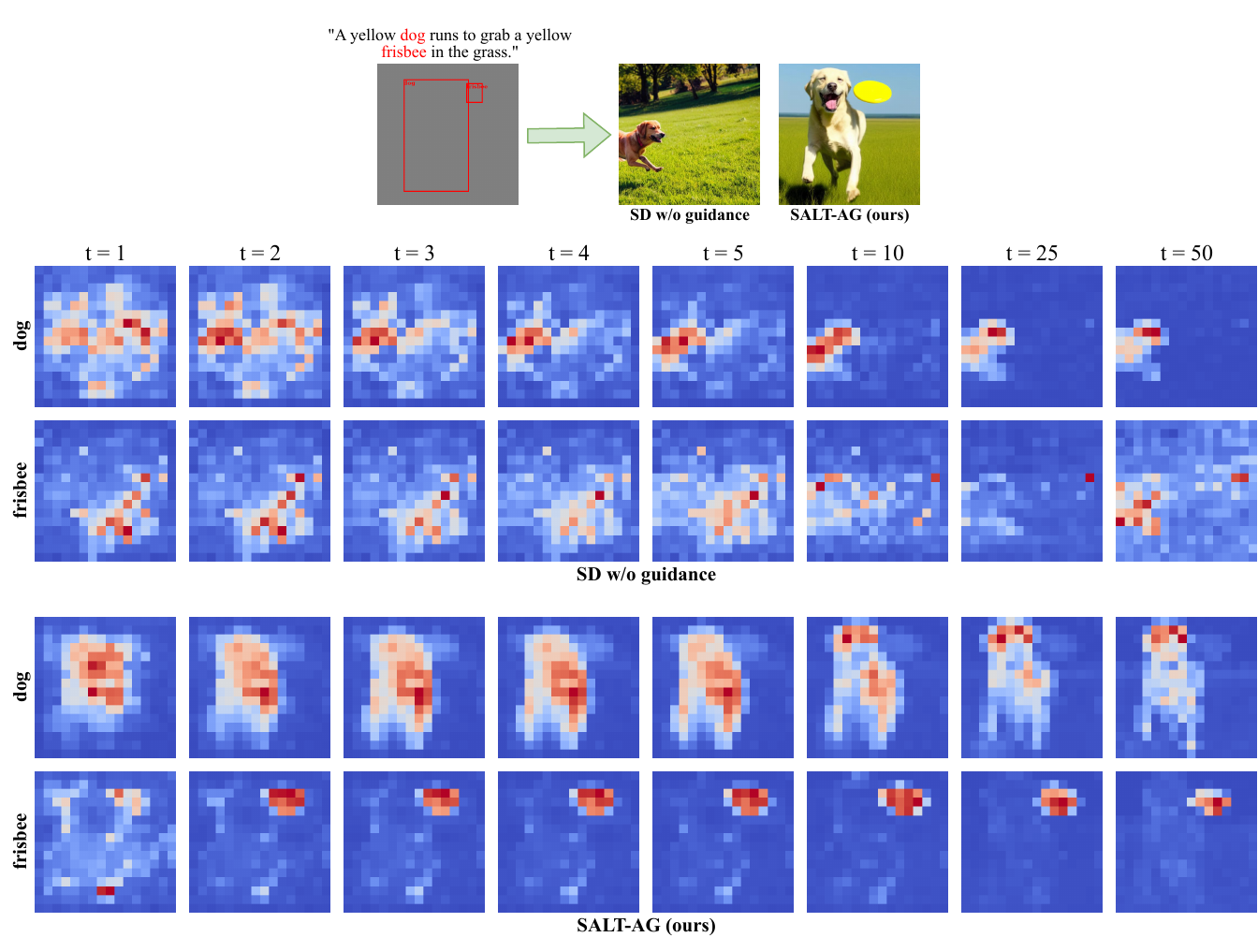}
        \caption{\textbf{Visualization of cross-attention maps during the denoising process.} Based on the given textual prompts and bounding boxes, we present the cross-attention maps of SD w/o guidance (second and third rows to the top) and our layout guidance approach (forth and fifth rows to the top) across different denoising steps.}
        \label{fig:append_atten_2}
\end{figure*}

\noindent \textbf{Cross-attention Maps Visualization.}\quad Given the textual prompts and bounding boxes, we visualize the cross-attention maps of different objects during the denoising process. As shown in Fig. \ref{fig:append_atten_1} and Fig. \ref{fig:append_atten_2}, compared with SD, the cross-attention maps in our approach converge to the given bounding box quickly.

\end{document}